\documentclass[letterpaper, 10 pt, conference]{ieeeconf}  

\IEEEoverridecommandlockouts                              

\usepackage{amsfonts} 
\usepackage{amsmath} 
\usepackage{booktabs}
\usepackage{tikz}
\usepackage{multirow}
\usepackage{threeparttable}
\usepackage{subcaption}
\usepackage[dvipsnames]{xcolor}
\usepackage{wrapfig}
\usepackage{hyperref}
\usepackage{balance}
\title{\LARGE \bf
Adversarial Vulnerabilities of Learned Telesurgery Policies
}

\author{Shutong Jin$^{1,2}$, Ziyang Chen$^{1}$, Preethi Satish$^{1}$, Paavan Gupta$^{1}$, Florian T. Pokorny$^{2}$ and Ken Goldberg$^{1}$
\thanks{$^{1}$Department of Electrical Engineering and Computer Sciences, University of California, Berkeley, CA 94720, USA.}%
\thanks{$^{2}$Department of Robotics, Perception and Learning, KTH Royal Institute of Technology, Stockholm, SE 10044, Sweden.}%
}

\begin{document}

\maketitle
\thispagestyle{empty}
\pagestyle{empty}

\begin{abstract}
While not yet in clinical deployment, learning-based policies are increasingly considered to augment the dexterity of human surgeons in robot-assisted surgery.
Can the end-to-end mapping from visual observations to robot actions be vulnerable to adversarial attacks?
We present the first study of adversarial vulnerabilities in learning-based policies for surgical robotics, conducted in a laboratory white-box setting where the attacker is assumed to have access to policy information and injects perturbations into the video stream transmitted over the network.
Two attack modes are considered: (a) disruptive attacks, where subtle visual perturbations interrupt policy execution without being noticed by a surgeon, and (b) steering attacks, where perturbations steer policy actions toward attacker-specified directions.
We study three adversarial attack methods, each with increasing access to policy information, and evaluate their impact on two surgical subtasks: debridement and suturing, performed on phantoms. 
Our evaluation covers three end-to-end policy architectures: ACT, Diffusion Policy, and $\pi_0$.
In addition, we identify a vulnerability to photometric perturbations, which mimic natural visual changes such as lighting variation.
Results from 620 physical experiments suggest that state-of-the-art policies can be significantly disrupted, resulting in an average 61\% reduction in surgical subtask success rates.
These findings suggest that adversarial vulnerabilities are important to consider for learned telesurgery policies.
Project page: 
\href{https://surgical-robotics.github.io/adversarial-vulnerability/}{https://surgical-robotics.github.io/adversarial-vulnerability/}
\end{abstract}

\section{Introduction}
In robot-assisted surgery, systems such as the da Vinci platform (Intuitive Surgical Inc., USA) today rely on human surgeon teleoperation.
The quality and consistency of the surgical outcome depend heavily on surgeon experience and performance, and surgery imposes substantial workload. 
Augmented dexterity, proposed by Goldberg and Guthart in 2024 \cite{goldberg2024augmented}, envisions robots that assist surgeons with repetitive tasks under close human supervision to alleviate surgeon burden and help standardize procedures.
Data-driven imitation learning policies using neural networks have been explored for augmenting dexterity in applications ranging from endoscope control~\cite{pore2023autonomous} to knot tying \cite{chen2025surgical} to general-purpose surgery~\cite{schmidgall2024general}.
As there are over 11,000 da Vinci systems installed worldwide and more than 20 million procedures performed by the end of 2025~\cite{intuitive2025report}, a large and growing body of expert-generated surgical data could enable learning-based policies for surgical augmented dexterity.

However, a long-standing risk associated with neural networks has received limited attention in the context of surgery.
In computer vision, adversarial attacks that create imperceptible visual perturbations to input images can cause Convolutional Neural Networks (CNNs) to misclassify~\cite{goodfellow2015explaining}.
This raises related questions for robotic surgery, where vision from endoscopes plays a central role: 

\textit{Are learned policies vulnerable to adversarial attacks in surgical manipulation tasks?
Can subtle perturbations in the camera images induce sudden and potentially unsafe robot motions?}

These questions are underexplored in surgical robotics, despite its safety-critical nature and direct implications for patient safety, yet the release of datasets, policy weights, and related implementation details may provide attackers with more information to design malicious attacks against policies applied to surgical tasks~\cite{nelson2026open}.
A growing trend in telesurgery~\cite{heemeyer2025telesurgery}, motivated by extending specialist care to remote regions and shortening time to treatment, additionally places the control loop over public or shared networks, exposing a further attack surface.

\begin{figure}[t]
    \centering
    \includegraphics[width=0.9\linewidth]{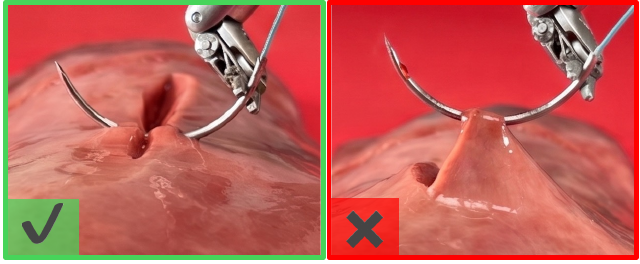} 
    \caption{Conceptual illustration of an adversarial attack on a suturing subtask, shown on ex vivo porcine liver. Left: execution with intended needle-tissue interaction. Right: steering attack that suddenly retracts the gripper, causing irreversible tissue deformation and potential scarring.}
    \label{fig:first_figure}
    \vspace{-1em}
\end{figure}

We present the first systematic study of adversarial vulnerabilities of learning-based policies for surgical augmented dexterity.
We consider two attack modes relevant to surgical robotics: 
(a) disruptive attacks, where subtle perturbations can interrupt policy execution; 
and (b) steering attacks, where perturbations can steer the policy toward an attacker-specified action direction, such as excessive needle lifting or penetration during suturing.
Under these two modes, we formulate three adversarial attacks with different levels of access to policy information, including policy observations, training data, and policy weights.
We then evaluate their impact on state-of-the-art policy architectures for augmented dexterity in robotic surgery, including Action Chunking Transformer (ACT)~\cite{zhao2023learning}, Diffusion Policy~\cite{chi2025diffusion}, and $\pi_0$~\cite{black2025pi0}, across two surgical subtasks: debridement and suturing.
This paper makes three contributions:
\begin{enumerate}
    \item We present the first study of adversarial vulnerabilities of learning-based surgical manipulation policies, identifying two attack modes and evaluating three attack methods for each mode.
    \item We identify a vulnerability to photometric perturbations, which mimic natural visual changes such as lighting variation while steering policy outputs toward attacker-specified directions.
    \item Results from 620 physical experiments suggesting that state-of-the-art policies are highly vulnerable to these attacks, with an average task success rate drop of 61\%. Steering attacks can be generated within milliseconds per observation and can amplify small actions into large unsafe actions.
\end{enumerate}

\section{Related Work}
\subsection{Augmented Dexterity in Surgical Robotics}
Augmented dexterity is a term proposed by Goldberg and Guthart \cite{goldberg2024augmented} as a more palatable alternative to ``supervised autonomy" in robotic surgery. Prior work commonly decomposes the augmented dexterity pipeline into perception, decision-making, and execution~\cite{fiorini2022concepts}. On the perception side, studies have focused on surgical scene segmentation,
instrument tracking,
and workflow recognition~\cite{chen2024toward}. 
These perceptual capabilities support high-level decision-making, where existing methods have explored task planning~\cite{pedram2017autonomous}, confidence-based autonomy allocation~\cite{kam2021confidence}, and hierarchical task selection~\cite{kim2025srt}. 
At the execution level, augmented dexterity systems have progressed from structured subtasks such as Fundamentals of Laparoscopic Surgery peg transfer, suturing, and knot tying~\cite{chen2025surgical, pedram2017autonomous} to deformable tissue manipulation and ex vivo anastomosis~\cite{shin2019autonomous, saeidi2022autonomous}.
More recently, learning-based policies are of increasing interest for integrating perception, decision-making, and execution in robot-assisted surgery. By learning from human demonstrations, these policies have been applied to surgical robotic manipulation subtasks such as tissue handling, needle manipulation, suturing, and knot tying~\cite{kim2024surgical}.

\subsection{Adversarial Attacks on CNNs}
Adversarial attacks refer to maliciously crafted perturbations $\delta$ added to an input image $x$ that cause a CNN $f$ to misclassify while keeping the perturbed image visually similar to the original~\cite{szegedy2014intriguing}:
\begin{equation}
    f(x+\delta) \neq y.
    \label{eq:cv_adversarial_attack}
\end{equation}
Here, $y$ denotes the true label.
Such perturbations are often obtained by following model-gradient directions that increase classification loss~\cite{goodfellow2015explaining}.
To control visibility, existing attacks commonly bound the perturbation magnitude~\cite{madry2018towards} or restrict the perturbation to a localized image patch~\cite{brown2017adversarial}.
The study of adversarial attacks has moved from image classification~\cite{moosavi2016deepfool} to broader perception tasks such as semantic segmentation
and object detection~\cite{wei2019transferable}.
Recent work has begun studying adversarial attacks on learned manipulation policies~\cite{chen2024diffusion,kalra2025vulnerable}.
Adversarial attacks on Vision-Language-Action (VLA) models have used adversarial patches
or 3D textures~\cite{chen2026tex3d}, which place visible crafted patterns in the scene. Language-based attacks instead inject malicious tokens into prompts~\cite{jones2025adversarial}, while action-freezing attacks halt the policy execution~\cite{wang2025freezevla}.
Previous work has examined visually conspicuous perturbations, which are less relevant to the augmented dexterity setting, where a surgeon supervises execution. 
In this work, we study the vulnerability of surgical policies to subtle changes in the visual stream, and further identify a vulnerability to photometric perturbations, which mimic natural visual changes such as lighting variation.

\subsection{Telesurgery}
Telesurgery separates the surgeon's console from the patient-side manipulators, connecting them over a network~\cite{heemeyer2025telesurgery}.
Growing interest is driven by the potential to reach regions with sparse surgical expertise and to shorten time to treatment in urgent cases.
Clinical feasibility was first demonstrated in 2001, when a surgeon in New York removed a patient's gallbladder in Strasbourg over a leased private telecommunication line~\cite{marescaux2001transatlantic}.
Since then, telesurgery has been reported across a range of procedures, including radical prostatectomy,
gastrectomy,
and percutaneous coronary intervention~\cite{patel2019long}.
Recent studies have moved from leased private circuits to commercial 5G and fiber infrastructure,
as leased circuits are too costly to scale and unavailable in rural regions~\cite{hara2025toward}.
While this broadens access, it also places the control loop on shared infrastructure, where prior work has demonstrated interception of surgeon commands through an unauthenticated teleoperation protocol~\cite{bonaci2015make}.
In this paper, we investigate the video stream transmitted over the network as an exposed channel, and characterize whether subtle perturbations injected into it can disrupt or steer a learned policy.

\section{Problem Formulation}
\label{sec:problem_formulation}

\paragraph{Learning from Surgical Demonstrations}
We investigate the vulnerability of learned surgical policies to adversarial perturbation, covering both vision-conditioned policies (ACT, Diffusion Policy) and VLA models ($\pi_0$).
We consider a two-stage imitation learning framework for learned policies.
During training, an expert demonstrator teleoperates the da Vinci robot through the console to collect a demonstration dataset $\mathcal{D}=\{(o,a)\}$, where $o=(i,p)\in\mathcal{O}$ contains the endoscopic camera image $i$ and robot proprioception $p$, and $a\in\mathcal{A}$ denotes the expert action.
We train a policy $\pi_\theta:\mathcal{O}\rightarrow\mathcal{A}$ to predict expert actions from observations, yielding a learned policy $\pi_{\theta}$.
During inference, the network weights $\theta$ are fixed. 
At each time step $t$, the policy maps $o_t$ to $a_t$, which are executed by the robot.

\begin{figure*}[t]
    \centering
    \includegraphics[width=0.75\linewidth]{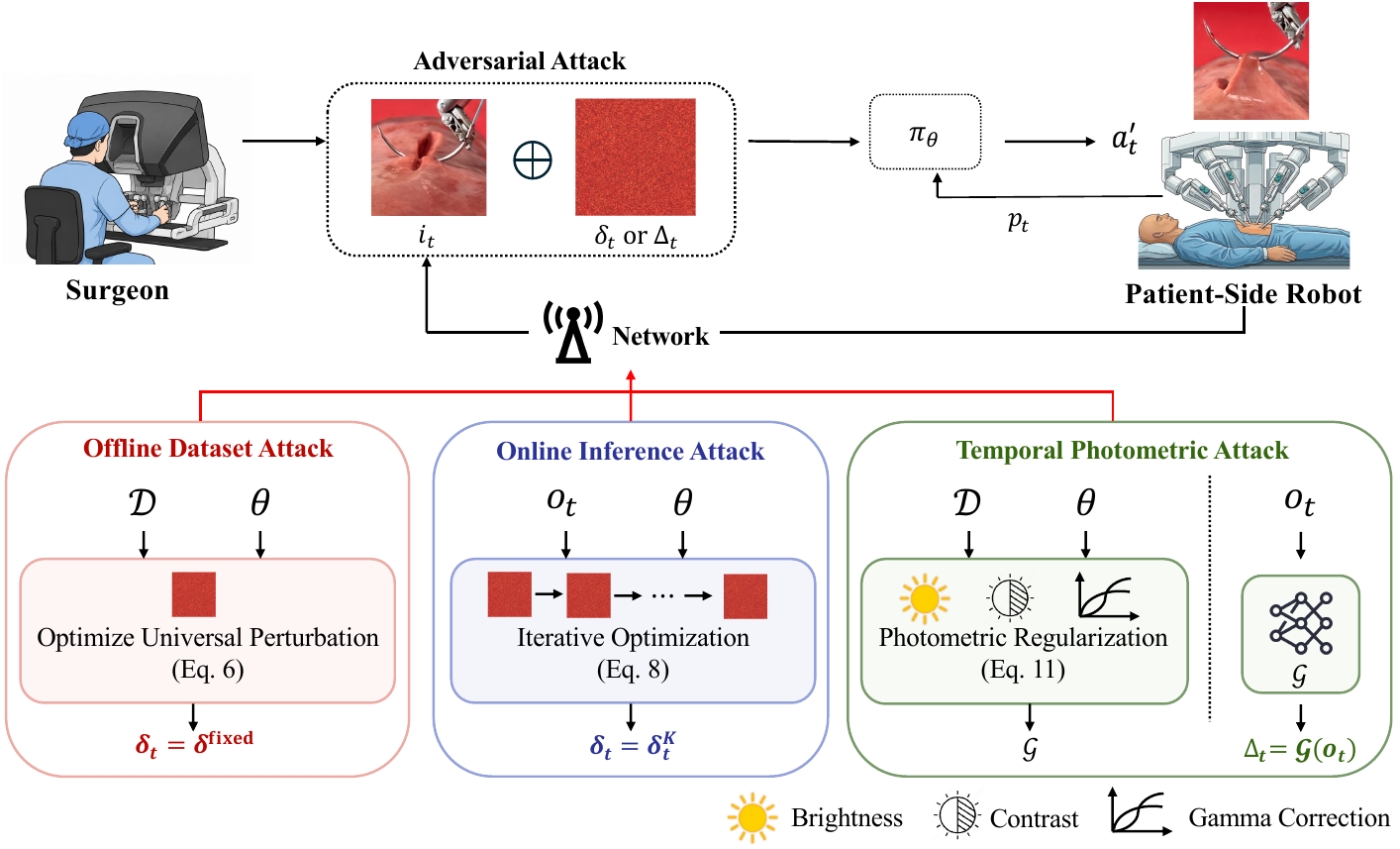}
    \caption{Illustration of the considered adversarial attack scenario. During an ongoing telesurgery at timestep $t$, the surgeon oversees the procedure via the live endoscopic video stream. An adversary injects either an imperceptible perturbation $\delta_t$ or a visually subtle photometric perturbation $\Delta_t$ into the clean endoscopic image $i_t$.
    The visually disguised input tricks the robot into executing an anomalous action $a_t'$ before the surgeon detects the anomaly. These perturbations are generated by the three attack methods discussed in \textit{Sec.}~\ref{sec:delta} under different levels of access to the training dataset $\mathcal{D}$, policy weights $\theta$, and current observation $o_t$.}
    \label{fig:pipeline}
    \vspace{-1em}
\end{figure*}

\paragraph{Adversarial Attacks}
As shown in \textit{Fig.}~\ref{fig:pipeline}, during telesurgery at timestep $t$, the adversary adds a visual perturbation to the endoscopic image $i_t$, producing an attacked image:
\begin{equation}
    i_t' = i_t + \delta_t,
    \qquad
    i_t' = i_t + \Delta_t,
    \label{eq:robot_adversarial_attack}
\end{equation}
where $\delta_t$ denotes an imperceptible visual perturbation bounded by a magnitude constraint, while $\Delta_t$ denotes a visually subtle perturbation designed to resemble natural visual variations.
The attacked image $i_t'$, together with clean robot proprioception $p_t$, is fed into the policy to produce a potentially unsafe action:
\begin{equation}
    a_t' = \pi_{\theta}(i_t', p_t).
\end{equation}

\paragraph{Attack Modes}
We consider two attack modes.
The disruptive attack interrupts the policy by maximizing the deviation from the clean action:
\begin{equation}
    \mathcal{L}_{\mathrm{disruptive}}
    =
    -
    \left\|
    a' - 
    a
    \right\|_2^2.
\end{equation}
The steering attack drives the policy output toward an attacker-specified direction:
\begin{equation}
    \mathcal{L}_{\mathrm{steering}}
    =
    \left\|
    a' - a^{\mathrm{target}}
    \right\|_2^2 .
\end{equation}
We define the target action as $a^{\mathrm{target}} = a + b$, where $b$ is an attacker-specified per-joint offset that specifies the desired action-space directional shift.

\paragraph{Assumptions}
We make the following assumptions: 
1) the attacks are white-box~\cite{biggio2018wild}, where the attacker has access to policy information, such as the training dataset $\mathcal{D}$, policy parameters $\theta$, and current observation $o_t$, with the exact access specified for each attack setting in \textit{Sec.}~\ref{sec:delta}; 
2) the attacker perturbs only the image component $i_t$ of the observation $o_t=(i_t,p_t)$ and does not issue direct robot commands or modify the controller.

\section{Attack Generation Methods}
\label{sec:delta}
This section investigates three adversarial perturbation generation methods under different levels of access to policy information.

\subsection{Offline Dataset Attack}
\label{sec:uap}
In the offline dataset attack, the adversary uses the training dataset $\mathcal{D}$ and the trained policy weights $\theta$ to compute a fixed visual perturbation before deployment, which is then reused during policy execution without online optimization.
We instantiate this setting with Universal Adversarial Perturbations (UAP)~\cite{moosavi2017universal}, where the fixed perturbation $\delta^{\mathrm{fixed}}$ is optimized with Adam~\cite{kingma2014adam} over all visual observations in $\mathcal{D}$:
\begin{equation}
    \delta^{\mathrm{fixed}}
    =
    \arg\min_{\delta}
    \frac{1}{|\mathcal{D}|}
    \sum^{\mathcal{D}}
    \mathcal{L}_{\mathrm{attack}} ,
    \label{eq:uap_attack}
\end{equation}
\begin{equation}
    \|\delta^{\mathrm{fixed}}\| \leq \epsilon ,
    \label{eq:uap_epsilon}
\end{equation}
where $\mathcal{L}_{\mathrm{attack}} \in \{\mathcal{L}_{\mathrm{disruptive}}, \mathcal{L}_{\mathrm{steering}}\}$ denotes one of the two attack modes defined in \textit{Sec.}~\ref{sec:problem_formulation}, and $\epsilon$ bounds the perturbation magnitude to preserve visual imperceptibility.
During online policy inference, the same fixed perturbation is applied to every visual observation, i.e., $\delta_t = \delta^{\mathrm{fixed}}$.

\subsection{Online Inference Attack}
\label{sec:pgd}
In the online inference attack, the adversary computes perturbations while the robot is operating.
At each timestep $t$, the adversary observes the current clean observation $o_t$ and uses the policy weights $\theta$ to compute an observation-specific perturbation.
Unlike the offline dataset attack, this perturbation is optimized independently for each timestep rather than reused across observations.
We instantiate this setting with Projected Gradient Descent (PGD)~\cite{madry2018towards}, which starts from an initial perturbation $\delta_t^0$ and iteratively optimizes it according to the selected attack mode:
\begin{equation}
    \delta_t^{k+1}
    =
    \delta_t^k
    -
    \alpha
    \nabla_{\delta_t^k}
    \mathcal{L}_{\mathrm{attack}},
    \label{eq:pgd_attack}
\end{equation}
\begin{equation}
    \|\delta_t^{k+1}\| \leq \epsilon,
    \qquad
    k=0,\ldots,K-1,
    \label{eq:pgd_epsilon}
\end{equation}
where $\nabla_{\delta_t^k}\mathcal{L}_{\mathrm{attack}}$ denotes the gradient of the attack loss with respect to the current perturbation $\delta_t^k$, $\alpha$ is the step size, and $K$ is the number of optimization steps.
For PGD, $\mathcal{L}_{\mathrm{attack}}$ is computed using the online clean action $a_t$ and attacked action $a_t'$, rather than offline dataset actions.
During online policy inference, we perform this optimization independently for each observation and apply the final-step perturbation, i.e., $\delta_t = \delta_t^K$.

\subsection{Temporal Photometric Attack}
\label{sec:tpa}
UAP applies one fixed perturbation across offline observations, which may limit its adaptation to unseen observations during closed-loop policy execution.
PGD optimizes a perturbation for each observation, which may incur higher computation.
We formulate Temporal Photometric Attack (TPA) as an intermediate approach:
it generates per-observation perturbations through a single forward pass of a trained generator.
TPA separates attack generation into offline training and online deployment.

\textbf{Offline Training. }
We use a lightweight convolutional network $\mathcal{G}$~\cite{baluja2017adversarial} to predict perturbations from clean observations.
During training, the attack assumes access to the dataset $\mathcal{D}$ and the policy weights $\theta$.
We train $\mathcal{G}$ using the following objective:
\begin{equation}
    \mathcal{L}
    =
    \mathcal{L}_{\mathrm{attack}}
    +
    \lambda_\mathrm{photo} \mathcal{L}_{\mathrm{photo}}.
\end{equation}
Here, $\lambda_\mathrm{photo}$ controls the strength of the photometric regularization term $\mathcal{L}_\mathrm{photo}$.
Instead of enforcing the hard imperceptibility constraint $\|\delta\| \leq \epsilon$ used in \textit{Eq.}~\ref{eq:uap_epsilon} and \textit{Eq.}~\ref{eq:pgd_epsilon}, we propose the photometric regularization term $\mathcal{L}_\mathrm{photo}$ to encourage photometrically plausible visual perturbations.
This relaxes the perturbation constraint while keeping the resulting changes aligned with natural photometric variations.
We define the regularization term as:
\begin{equation}
    \mathcal{L}_{\mathrm{photo}}
    =
    \left\|
    i + \Delta
    -
    \mathcal{T}_{\mathrm{photo}}(i)
    \right\|_2^2,
\end{equation}
where $\mathcal{T}_{\mathrm{photo}}$ denotes a visually plausible photometric transformation, such as brightness, contrast, or gamma adjustment.
Details and examples are provided in the \hyperref[appendix]{Appendix}.

\textbf{Online Deployment. }
During online inference, the generator predicts an adversarial perturbation for each clean observation $o_t$ in a single forward pass:
\begin{equation}
    \Delta_t = \mathcal{G}(o_t).
\end{equation}

\section{Experiments}
\label{sec:experiments}
\subsection{Implementation Details}
\paragraph{Experimental Tasks}
We consider two surgical subtasks that require distinct motion primitives: debridement, which involves the removal of infected or dead tissue fragments from a wound, and suturing, a delicate procedure for secure wound closure via surgical needle manipulation. In our experimental setup for debridement, a deformable cubic foam is randomly placed on a flesh-like soft phantom, and the da Vinci Research Kit (dVRK) gripper is controlled to grasp the foam fragment and transfer it into a container. For the suturing subtask, the gripper manipulates a surgical needle to insert it into a soft tissue phantom featuring a raised wound.

\paragraph{Policy Training}
We train policies for surgical manipulation tasks using three widely used, high-performing architectures: ACT~\cite{zhao2023learning}, Diffusion Policy~\cite{chi2025diffusion}, and $\pi_0$~\cite{black2025pi0}. 
During data collection, an experienced operator performed 80 trials of debridement and 100 trials of suturing.
Each policy takes a third-view RGB image from a mounted Allied Vision camera with a resolution of ${224}\times{224}$ and the dVRK Patient-Side Manipulator (PSM) proprioception as input.
The action space follows the 7-DoF PSM kinematic structure: arm yaw/pitch/insertion, wrist roll/pitch/yaw, and jaw actuation.

\paragraph{Evaluation Protocols}
We evaluate policy vulnerability and attack efficacy from four perspectives: 1) \textit{Task Success Rate}: The fraction of trials in which the policy completes the task under adversarial attack. 
2) \textit{Time Efficiency}: The average wall-clock time required to generate and load one attack to one observation during online policy inference.
3) \textit{Attack Visual Similarity}: The visual similarity between clean frames and attacked frames. We use SSIM~\cite{wang2004image} to measure frame-level structural similarity and temporal SSIM~\cite{wang2004video} to measure structural consistency between consecutive attacked frames.
4) \textit{Attack Strength}: The attack-induced change on target joint $j$ along the attacker-specified direction, normalized as a \% of the original action on that joint.
Positive/negative values indicate shifts with/against the specified direction.
We report relative change because absolute deviation depends on episode length, which varies across subtasks and with operator intervention.
Each combination of policy architecture, attack method, and attack mode is evaluated over 10 trials, yielding 620 real-world evaluations in total.

\paragraph{Questions}
We investigate the following questions:
\begin{enumerate}
    \item How vulnerable are learned policies to adversarial attacks in surgical manipulation tasks?  
    \item Can steering attacks induce action shifts in the attacker-specified direction? 
    \item How perceptible are the attacks, as measured by image similarity metrics and human detection latency?
    \item Do the attacks remain effective across policies, tasks, and defenses?
\end{enumerate}

\begin{table*}[t]
    \centering
    \caption{Debridement performance after UAP (Offline Dataset Attack), PGD (Online Inference Attack), and TPA (Temporal Photometric Attack). Bold indicates the best performance, and underline indicates the second best. The $\text{Clean}$ row reports policy execution under clean observations without perturbations.}
    \label{tab:attack_debridement}

    \footnotesize
    \setlength{\tabcolsep}{4pt}
    \renewcommand{\arraystretch}{1.0}

    \resizebox{\textwidth}{!}{%
    
    \begin{tabular}{@{}ll *{3}{r} @{\hspace{4pt}\vrule\hspace{4pt}} *{3}{r} @{\hspace{4pt}\vrule\hspace{4pt}} *{3}{r} @{\hspace{4pt}\vrule\hspace{4pt}} *{3}{r} @{\hspace{4pt}\vrule\hspace{4pt}} *{3}{r} @{}}
        \toprule

        & & \multicolumn{3}{c}{Task Success Rate ($\downarrow$)} 
          & \multicolumn{3}{c}{Time (ms, $\downarrow$)} 
          & \multicolumn{3}{c}{SSIM ($\uparrow$)} 
          & \multicolumn{3}{c}{Temporal SSIM ($\uparrow$)} 

          & \multicolumn{3}{c}{\shortstack{Attack Strength (\%, $\uparrow$)}} \\
        \cmidrule(lr){3-5} \cmidrule(lr){6-8} \cmidrule(lr){9-11} \cmidrule(lr){12-14} \cmidrule(lr){15-17}

        Objective & Method
        
        & \makebox[2.5em][r]{ACT} & \makebox[3.1em][r]{Diff.} & \makebox[2.5em][r]{$\pi_0$}
        & \makebox[2.5em][r]{ACT} & \makebox[2.5em][r]{Diff.} & \makebox[2.5em][r]{$\pi_0$}
        & \makebox[2.5em][r]{ACT} & \makebox[2.5em][r]{Diff.} & \makebox[2.5em][r]{$\pi_0$}
        & \makebox[2.5em][r]{ACT} & \makebox[2.5em][r]{Diff.} & \makebox[2.5em][r]{$\pi_0$}
        & \makebox[2.5em][r]{ACT} & \makebox[3.2em][r]{Diff.} & \makebox[2.5em][r]{$\pi_0$} \\
        \midrule      

        Clean & -- & 0.9 & 0.9 & 0.7 & -- & -- & -- & -- & -- & -- & -- & -- & -- & -- & -- & -- \\ 
        
        \midrule
        
        \multirow{4}{*}{Disruptive}
        & Random & 0.9 & 0.8 & \underline{0.6}  & \textbf{0.1} & \textbf{0.1} & \textbf{0.1}  & \textbf{0.97} & \textbf{0.97} & \underline{0.96} & \underline{0.989} & \textbf{0.989} & \underline{0.992}  & -- & -- & -- \\
        & UAP    & \underline{0.6} & \underline{0.5} & \textbf{0.0}  & \textbf{0.1} & \textbf{0.1} & \textbf{0.1}  & \underline{0.95} & 0.95 & 0.94  & \textbf{0.990} & \textbf{0.989} & \textbf{0.993}  & -- & -- & -- \\
        & PGD    & \textbf{0.0}   & \textbf{0.0}   & \textbf{0.0}  & 84.3 & 4642 & 3050 & \textbf{0.97} & \underline{0.96} & \textbf{0.97}  & 0.951 & 0.936 & 0.991  & -- & -- & -- \\
        & TPA    & \textbf{0.0}   & \textbf{0.0}   & \textbf{0.0}  & \underline{3.9} & \underline{7.4} & \underline{7.8}   & 0.81 & 0.68 & 0.76  & 0.987 & \underline{0.986} & 0.987  & -- & -- & -- \\

        \midrule

        \multirow{3}{*}{Steering}
        & UAP    & 0.8 & \underline{0.5} & \underline{0.3}  & \textbf{0.1} & \textbf{0.1} & \textbf{0.1}  & \underline{0.96} & \underline{0.94} & \underline{0.94}  & \textbf{0.989} & \textbf{0.990} & \textbf{0.993}  & 0.2 & \underline{0.7} & \underline{0.7} \\
        & PGD    & \underline{0.2} & \textbf{0.0}   & \textbf{0.0}  & 87.2 & 5024 & 3314 & \textbf{0.97} & \textbf{0.96} & \textbf{0.97}  & 0.961 & 0.963 &  0.938 & \underline{0.8} & -1.1 & -0.6 \\
        & TPA    & \textbf{0.1} & \textbf{0.0}   & \textbf{0.0}  & \underline{3.6} & \underline{13.5} & \underline{6.6}   & 0.90 & 0.83 & 0.88  & \underline{0.988} & \underline{0.986} & \underline{0.988}  & \textbf{4.6} & \textbf{1.9} & \textbf{3.1} \\

        \bottomrule
    \end{tabular}%
    }
\end{table*}
\begin{table*}[t]
    \centering
    \caption{Suturing performance after UAP (Offline Dataset Attack), PGD (Online Inference Attack), and TPA (Temporal Photometric Attack). }
    \label{tab:attack_suturing}

    \footnotesize
    \setlength{\tabcolsep}{4pt}
    \renewcommand{\arraystretch}{1.0}

    \resizebox{\textwidth}{!}{%

    \begin{tabular}{@{}ll *{3}{r} @{\hspace{4pt}\vrule\hspace{4pt}} *{3}{r} @{\hspace{4pt}\vrule\hspace{4pt}} *{3}{r} @{\hspace{4pt}\vrule\hspace{4pt}} *{3}{r} @{\hspace{4pt}\vrule\hspace{4pt}} *{3}{r} @{}}
        \toprule

        & & \multicolumn{3}{c}{Task Success Rate ($\downarrow$)} 
          & \multicolumn{3}{c}{Time (ms, $\downarrow$)} 
          & \multicolumn{3}{c}{SSIM ($\uparrow$)} 
          & \multicolumn{3}{c}{Temporal SSIM ($\uparrow$)} 
          & \multicolumn{3}{c}{\shortstack{Attack Strength (\%, $\uparrow$)}} \\
        \cmidrule(lr){3-5} \cmidrule(lr){6-8} \cmidrule(lr){9-11} \cmidrule(lr){12-14} \cmidrule(lr){15-17}

        Objective & Method

        & \makebox[2.5em][r]{ACT} & \makebox[3.1em][r]{Diff.} & \makebox[2.5em][r]{$\pi_0$}
        & \makebox[2.5em][r]{ACT} & \makebox[2.5em][r]{Diff.} & \makebox[2.5em][r]{$\pi_0$}
        & \makebox[2.5em][r]{ACT} & \makebox[2.5em][r]{Diff.} & \makebox[2.5em][r]{$\pi_0$}
        & \makebox[2.5em][r]{ACT} & \makebox[2.5em][r]{Diff.} & \makebox[2.5em][r]{$\pi_0$}
        & \makebox[2.5em][r]{ACT} & \makebox[3.2em][r]{Diff.} & \makebox[2.5em][r]{$\pi_0$} \\
        \midrule      

        Clean & -- & 0.6 & 0.7 & 0.5 & -- & -- & -- & -- & -- & -- & -- & -- & -- & -- & -- & -- \\ 
        
        \midrule
        
        \multirow{4}{*}{Disruptive}
        & Random & 0.5& 0.7 & 0.5& \textbf{0.1}& \textbf{0.1}& \textbf{0.1}& \underline{0.95} & \underline{0.95}& \underline{0.94}& \underline{0.987}& \underline{0.987}& \underline{0.990}&-- &-- &-- \\
        & UAP    & \underline{0.1} & \underline{0.3} & \underline{0.2}& \textbf{0.1}& \textbf{0.1}& \textbf{0.1}& \textbf{0.97}& \textbf{0.97}& \textbf{0.97}& \underline{0.987}& \textbf{0.988}& \textbf{0.991}&-- &-- &-- \\
        & PGD    & \textbf{0.0}& \textbf{0.0} & \textbf{0.0}& 103.3& 4841& 3201& 0.91 & 0.91& \underline{0.94}& 0.934& 0.925& 0.897&-- &-- &-- \\
        & TPA    & \textbf{0.0}& \textbf{0.0}& \textbf{0.0}& \underline{4.2}& \underline{17.8}& \underline{7.7}& \underline{0.95}& 0.94 & 0.81& \textbf{0.994}& \underline{0.987}& 0.989&-- &-- &-- \\

        \midrule

        \multirow{3}{*}{Steering}
        & UAP    &\textbf{0.0} & \underline{0.3} & \underline{0.1} & \textbf{0.1} & \textbf{0.1}& \textbf{0.1}& 0.90& \underline{0.91}& \underline{0.91}& \textbf{0.987}& \textbf{0.988}& \textbf{0.991}&0.1 & \underline{0.0}& \underline{0.1}\\
        & PGD    &\textbf{0.0} & \textbf{0.0} & \textbf{0.0}& 110.8& 5331& 3332& \underline{0.91}& \underline{0.91}& \textbf{0.94}& 0.939& 0.939& 0.902&\underline{0.2} &-0.1 &-0.3 \\
        & TPA    &\textbf{0.0} & \textbf{0.0} &\textbf{0.0} & \underline{3.9}& \underline{19.7}& \underline{7.8}& \textbf{0.95}& \textbf{0.95}& 0.85& \underline{0.986}& \underline{0.987}& \underline{0.989}& \textbf{0.7}& \textbf{0.2}& \textbf{0.3}\\

        \bottomrule
    \end{tabular}%
    }
\end{table*}
\begin{figure*}[h]
    \centering
    \includegraphics[width=\linewidth]{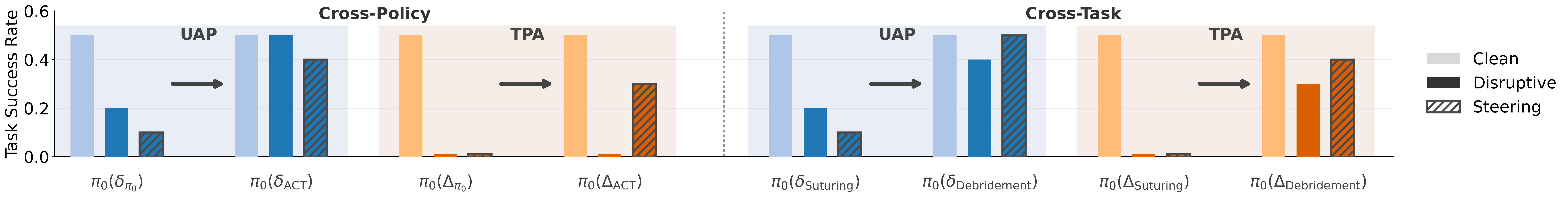}
    \caption{Cross-policy and cross-task transfer results for UAP and TPA. Each bar group reports three task success rates: clean execution, disruptive attack, and steering attack. For each arrow-connected pair, the left group shows performance under the original attack, and the right group shows performance using the transferred attack. $\pi_0(\delta_{\mathrm{ACT}})$ in the first arrow-connected pair denotes evaluating $\pi_0$ under a UAP attack generated from ACT; other labels follow the same convention.}
    \label{fig:cross_policy}
    \vspace{-0.5em}
\end{figure*}

\subsection{Results}

\paragraph{Evaluation of Policy Vulnerability under Disruptive Adversarial Attacks}
\label{sec:experiment_disruptive}
We evaluate disruptive attacks on two subtasks across three attack methods, using random imperceptible noise ($\text{Random}$) as a baseline to test whether visual perturbations alone cause performance degradation.
Results are reported in \textit{Tab.}~\ref{tab:attack_debridement} and \textit{Tab.}~\ref{tab:attack_suturing}.
Across all policy architectures, disruptive attacks substantially reduce task success rates, with average success-rate drop of 63\% for ACT, 67\% for Diffusion Policy, and 67\% for $\pi_0$ across two surgical subtasks. 
Disruptive attacks induce different failure behaviors across policies.
ACT and $\pi_0$ tend to show early gripper overshoot, where the dVRK gripper moves abruptly far beyond the intended motion direction.
Diffusion Policy produces smoother trajectories, but attack-induced errors accumulate over time and can cause misgrasping in debridement or suboptimal stitching poses in suturing, which may lead to needle snapping.
The random-noise baseline has little effect on all three policies, possibly due to visual-backbone pretraining and real-world sensory noise in the training data.
Suturing shows higher adversarial vulnerability than debridement because small directional errors can misalign needle insertion, causing bottom-surface collision, one-sided phantom penetration, or needle snapping.

\begin{figure*}[!t]
    \centering
    \includegraphics[width=0.714\linewidth]{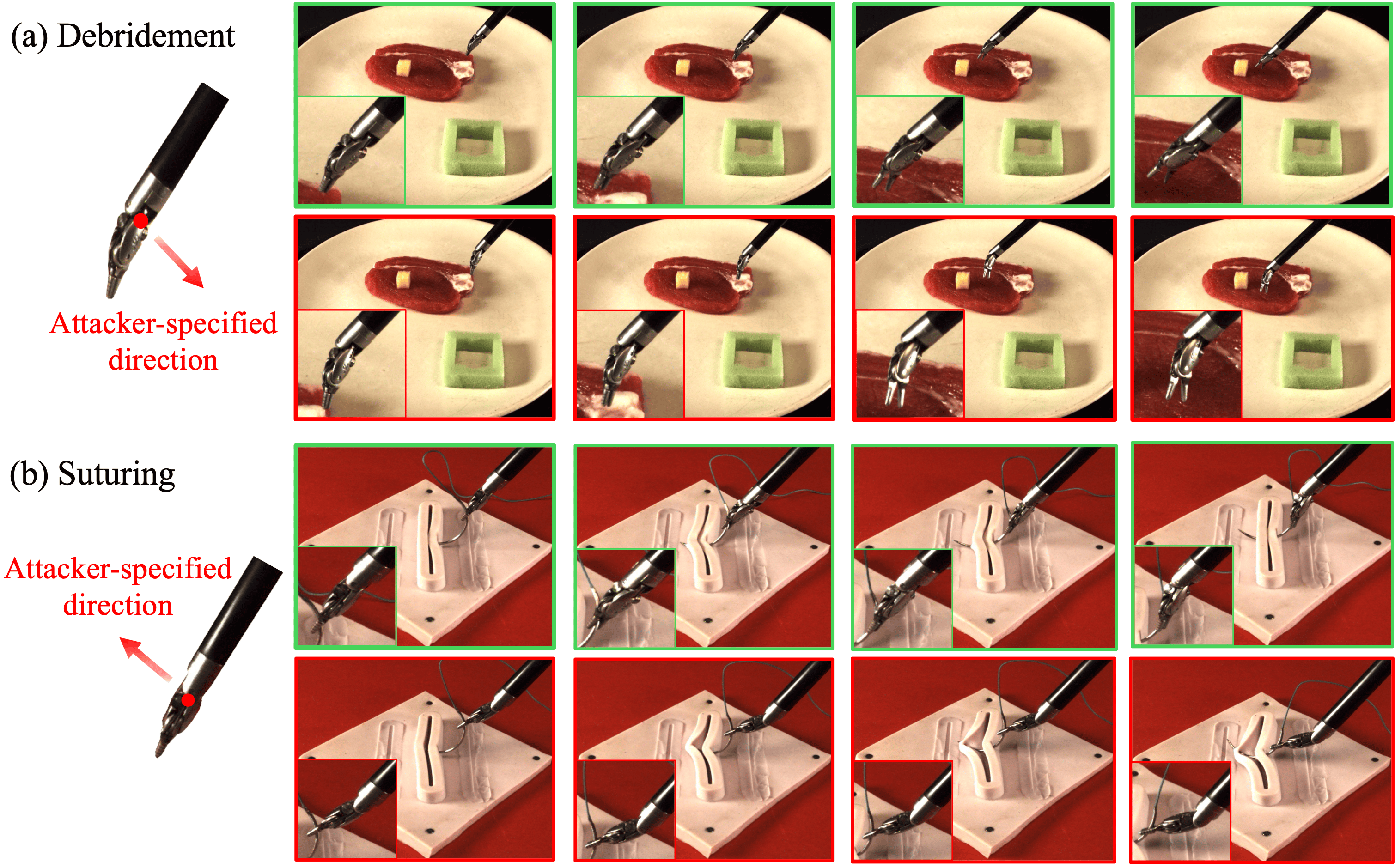} 
    \caption{Illustration of TPA steering attacks on (a) debridement and (b) suturing subtasks. For each subtask, the top row outlined in green shows policy execution under clean images, and the bottom row outlined in red shows policy execution under the TPA steering attack. For debridement, the attack targets a positive directional shift of the PSM wrist pitch joint (j4); for suturing, it targets a negative directional shift of the same joint. }
    \label{fig:keyframes_tasks}
    \vspace{-0.5em}
\end{figure*}

\paragraph{Evaluation of Steering Attack Effectiveness}
\label{sec:experiment_targeted}
We evaluate steering attacks generated by the three attack methods using task success rate and attack strength (\%), which quantifies the relative change on the target joint.
As shown in \textit{Tab.}~\ref{tab:attack_debridement} and \textit{Tab.}~\ref{tab:attack_suturing}, steering attacks reduce task success rates across policies.
UAP and PGD often show relatively weak steering performance: their attack strength values remain close to zero or slightly negative, indicating a failure to steer towards the attacker-specified direction. 
TPA shows stronger steering performance by replacing the hard imperceptibility constraint with photometric regularization, which allows a broader class of visually plausible perturbations.
We also observe that TPA reaches a higher average per-step attack strength in debridement than in suturing.
This difference is likely related to contact conditions: debridement leaves the attacked wrist joint mostly unconstrained, while suturing introduces resistance from the phantom.
As shown in \textit{Fig.}~\ref{fig:keyframes_tasks}, because robot control is sequential, these per-step steering biases can accumulate over time and gradually shift the trajectory toward the attacker-specified direction.

\paragraph{Evaluation of Visual Perceptibility}
\label{sec:experiment_imperceptible}
We evaluate the perceptibility of the crafted attacks from three perspectives: time efficiency, visual similarity, and human detection latency.
UAP is the fastest among the three attack generation methods because it reuses a fixed perturbation across images.
Although PGD often poses a stronger drop in task success rate than UAP, its optimization time is substantially longer and increases with model size; for example, attacking Diffusion Policy takes 5024 ms for a single observation.
TPA provides a balanced trade-off between attack effectiveness and generation time.
For visual perceptibility, UAP and PGD are visually similar under the visual similarity metrics, while TPA obtains lower SSIM scores.
This lower SSIM may result from the global affine photometric transformation used in TPA: although structural content is preserved, global tone changes can reduce image similarity.
For temporal consistency, TPA shows high temporal SSIM, ranking below only the temporally static random-noise baseline and UAP.
We additionally conduct a detection latency study on the debridement subtask to measure how long an attack remains undetected under human supervision.
In an informal experiment, nine students from our lab who were not involved in this project viewed recordings from the teleoperation console of the debridement subtask performed under UAP, PGD, and TPA disruptive attacks, and were instructed to press a key whenever they noticed anomalous behavior.
From the 27 trials, we report two metrics: 
(a) detection latency (s), the time from attack onset to the participant's keypress; 
and (b) workspace deviation (mm), the deviation of the attacked trajectory from a clean reference obtained by re-initializing the scene to the same initial configuration and running the policy without attack.
As shown in Tab.~\ref{tab:human}, mean detection latency ranges from 1.8 s (PGD) to 6.3 s (UAP), with corresponding workspace deviations of 11.7 mm (UAP) to 43.4 mm (TPA). 
Delays and deviations of this magnitude could be highly consequential in surgical manipulation, where precision is critical. 
PGD is detected fastest, as its per-observation perturbation is computed online and introduces a visible slowdown in the video stream.
TPA and UAP are detected later, with presses tending to occur after the gripper drifts off the taskboard or misses a grasp.
\begin{table}[h]
\centering
\caption{Human detection latency and workspace deviation under UAP, PGD and TPA attacks.}
\label{tab:human}
\begin{tabular}{lrrr}
\toprule
Metric & UAP & PGD & TPA \\
\midrule
Latency (s)    & $6.3 \pm 1.2$ & $1.8 \pm 0.6$ & $5.2 \pm 1.1$ \\
Deviation (mm) & $11.7 \pm 2.0$ & $12.1 \pm 1.4$ & $43.4 \pm 1.2$  \\
\bottomrule
\end{tabular}
\end{table}

\paragraph{Evaluation of Attack Effectiveness across Policies, Tasks and Defenses}
\label{sec:experiment_generalisability}
We evaluate attack transfer across policy architectures and tasks.
Specifically, we evaluate two transfer settings: cross-policy transfer, where attacks generated using ACT weights are applied to $\pi_0$ on suturing, and cross-task transfer, where attacks generated from the debridement dataset are applied to suturing using $\pi_0$.
We focus on UAP and TPA because PGD optimizes perturbations online for each observation and does not learn a reusable perturbation from the training dataset.
As shown in \textit{Fig.}~\ref{fig:cross_policy}, UAP transfers poorly across both policy architectures and tasks, likely because its fixed perturbation cannot adapt to the unseen observations.
TPA transfers across policy architectures under the disruptive mode but shows limited cross-task transfer.
This is likely because TPA learns its attack generator from the policy training dataset, which can remain relevant across architectures trained on the same task but may not match the visual and action distributions of a different task.
We additionally investigate whether these attacks can be mitigated by widely used image-space defenses. 
Two representative defenses with different deployment protocols are evaluated: (1) JPEG compression~\cite{guo2017countering}, which reduces the high-frequency content that adversarial perturbations rely on and is applied at inference time without training; and (2) adversarial training in the TRADES formulation~\cite{zhang2019theoretically}, which penalizes action deviation under worst-case perturbations recomputed at every policy training step. 
Training alternates per batch among the disruptive variants of UAP, PGD, and TPA, and inference is unchanged.
All experiments use ACT on the debridement subtask, with 10 trials per condition. 
JPEG compression, applied with quality factor $Q=75$ as in the original work, mitigates UAP almost fully and PGD partially. Success rates recover from 0.6 to 0.8 under UAP and from 0 to 0.3 under PGD, while TPA is unaffected, remaining at 0.
This may be because the norm constraint confines UAP and PGD to small per-pixel changes, which quantization removes, while TPA applies image-wide photometric changes that compression preserves.
Adversarial training instead incurs a substantial utility cost. 
The trade-off parameter was swept over $\beta \in \{1, 3, 6\}$, where $6$ is the value used in the original work, and set to $\beta = 1$: larger values degraded clean performance further.
The TRADES-trained policy still reaches only 0.3 without attack, down from 0.9, with failures dominated by imprecise grasping.
Under attack, the policy reaches 0.3 for UAP, 0.2 for PGD, and 0 for TPA.
TPA remains fully effective because its generator adapts to each observation rather than converging to a fixed perturbation, whereas UAP is a single fixed perturbation and is therefore fully covered by the training distribution.

\section{Limitation and Conclusion}
\label{sec:conclusion}
\paragraph{Limitations and Future Work}
This paper studies adversarial attacks on learning-based policies in surgical manipulation tasks, focusing on visual inputs.
Since robot are inherently multimodal, future work should examine attacks on other input modalities, e.g., force, and their interactions with visual perturbations.
We assume a white-box setting with access to the policy weights, which is consistent with the current trend toward openly released policies. The study therefore remains theoretical, with black-box conditions covered only by the cross-policy and cross-task transfer experiments. A systematic treatment of restricted-access attacks remains future work.
Additionally, this paper proposes Temporal Photometric Attack, a class of photometric adversarial attacks that disguise perturbations as natural visual variations.
Since this formulation is not specific to surgical robotics, future work can evaluate its impact in other safety-critical real-world systems, such as assistive manipulation or autonomous navigation.

\paragraph{Conclusion}
In this paper, we present the first study of adversarial vulnerabilities of learning-based policies for surgical manipulation tasks.
We investigate two attack modes, disruptive and steering, and evaluate three attack methods with different levels of policy information access under each mode.
We further identify a vulnerability to photometric perturbations, characterized through Temporal Photometric Attack (TPA), in which perturbations mimic natural visual changes such as lighting variation steer policies toward attacker-specified directions.
Through 620 physical experiments on two surgical subtasks, our results show that state-of-the-art policies are highly vulnerable to adversarial attacks, with an average success-rate drop of 61\%.
We also find that steering attacks can be generated within milliseconds per observation, amplifying small actions into large unsafe actions through closed-loop execution.
A human latency study further shows that observers take seconds to notice an ongoing attack, while classical defenses such as JPEG compression and adversarial training offer only partial mitigation, the latter at a substantial cost to clean success.

\paragraph{Responsible Disclosure}
We investigate the theoretical vulnerability of several of today's state-of-the-art manipulation policies to visual perturbations prior to any clinical deployment of the technology. Our work serves primarily to highlight potential safety challenges that the research community needs to address and mitigate prior to the deployment of these approaches to real-world clinical settings. All experiments are conducted in a laboratory setting on phantoms.

\section{Appendix}
\label{appendix}
We consider three photometric transformations $\mathcal{T}_{\mathrm{photo}}$ (\textit{Tab.}~\ref{tab:photo}): brightness variation, contrast variation, and gamma correction variation.
On every forward pass, the parameters of each transformation are fitted to the clean-attacked pair $(i,i')$ by least squares.
Here, $\mathbb{E}[\cdot]$, $\mathrm{Cov}(\cdot,\cdot)$, and $\mathrm{Var}(\cdot)$ denote the mean, covariance, and variance over all pixels in a batch.
They are computed per channel for brightness and contrast ($k,q\in\mathbb{R}^{3}$), and jointly over channels for gamma ($r\in\mathbb{R}$).
We set $\eta=10^{-8}$ for numerical stability.
All transformations are applied pixel-wise, with outputs clipped to the valid image range.
\begin{table}[h]
\centering
\caption{The three photometric transformations $\mathcal{T}_{\mathrm{photo}}$ and their least-squares parameter estimators.}
\label{tab:photo}
\setlength{\tabcolsep}{4pt}
\begin{tabular}{@{}llc@{}}
\toprule
Variation & $\mathcal{T}_{\mathrm{photo}}$ & Least-squares estimate \\
\midrule
Brightness & $k\,i$
  & $k = \dfrac{\mathbb{E}[i i']}{\mathbb{E}[i^2]+\eta}$ \\[2.2ex]
Contrast & $k\,i + q$
  & $k = \dfrac{\mathrm{Cov}(i,i')}{\mathrm{Var}(i)+\eta},\quad
      q = \mathbb{E}[i'] - k\,\mathbb{E}[i]$ \\[2.2ex]
Gamma & $i^{\,r}$
  & $r = \dfrac{\mathbb{E}[\log i\,\log i']}{\mathbb{E}[(\log i)^2]+\eta}$ \\
\bottomrule
\end{tabular}
\vspace{-1em}
\end{table}
\begin{figure}[h]
    \centering
    \begin{subfigure}[t]{0.3\linewidth}
        \centering
        \includegraphics[width=\linewidth]{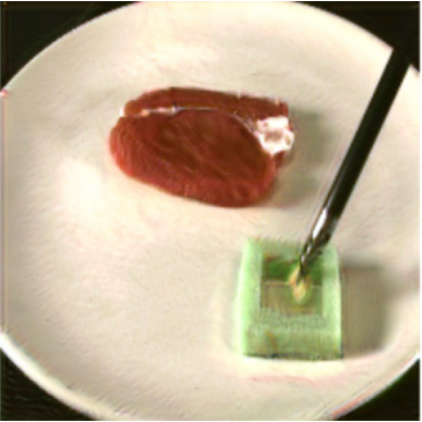}
        \caption{Brightness}
        \label{fig:tpa_brightness}
    \end{subfigure}
    \begin{subfigure}[t]{0.3\linewidth}
        \centering
        \includegraphics[width=\linewidth]{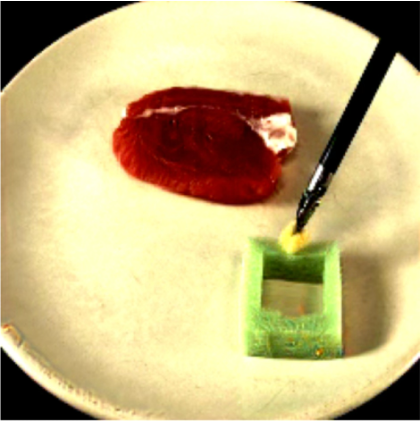}
        \caption{Contrast}
        \label{fig:tpa_contrast}
    \end{subfigure}
    \begin{subfigure}[t]{0.3\linewidth}
        \centering
        \includegraphics[width=\linewidth]{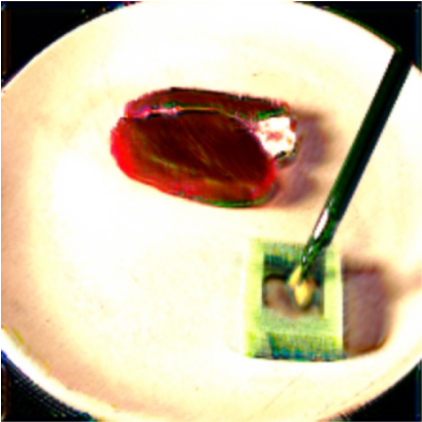}
        \caption{Gamma}
        \label{fig:tpa_gamma}
    \end{subfigure}
    \caption{Examples of attacked images generated with three photometric transformations.}
    \label{fig:tpa_regularization_examples}
\end{figure}

\section*{ACKNOWLEDGMENT}
This work was partially supported by the Wallenberg AI, Autonomous Systems and Software Program (WASP) funded by the Knut and Alice Wallenberg Foundation. The computations were enabled by the supercomputing resource Berzelius provided by the National Supercomputer Centre at Linköping University and the Knut and Alice Wallenberg Foundation, Sweden.

\balance
\bibliographystyle{IEEEtran}
\bibliography{citations}
\end{document}